\documentclass[11pt,oneside,a4paper]{article}
\pdfoutput=1
\usepackage{authblk}
\usepackage{mathptmx}
\usepackage{hyperref}
\usepackage{graphicx}
\usepackage{amsmath}
\usepackage{amssymb}
\usepackage{lmodern}
\usepackage{color}
\usepackage{todonotes}
\usepackage{booktabs}
\usepackage{wrapfig}
\usepackage[font={small},labelfont=bf]{caption}
\usepackage[top=1.2in, bottom=1.4in, left=1.3in, right=1.3in]{geometry}

\newif\ifnofigures

\newcommand{\challenge}{\textit{GlaS@MICCAI2015}}

\DeclareMathOperator*{\argmax}{arg\,max}

\newcommand{\textIf}[1]{\text{if} \quad #1}
\newcommand{\R}{\mathbb{R}}

\newcommand{\norm}[1]{\lVert #1 \rVert}
\newcommand{\abs}[1]{\lvert #1 \rvert}
\newcommand{\grad}{\nabla}

\newcommand{\TVNormCont}[1]{\abs{\grad #1}}

\newcommand{\dd}[1]{\,{\rm d} #1}
\newcommand{\intOmega}{\int_{\Omega}}

\newcommand{\MIN}[1]{\min\limits_{#1}}

\newcommand{\suchThat}{\qquad \text{s.t.}\qquad}

\begin{document}
\title{\bfseries Semantic Segmentation of Colon Glands with Deep Convolutional Neural Networks and Total Variation Segmentation}

\author[1,2,*]{Philipp Kainz}
\author[2]{Michael Pfeiffer}
\author[3,4,5]{Martin Urschler}
\affil[1]{Institute of Biophysics, Center for Physiological Medicine, Medical University of Graz, Graz, Austria}
\affil[2]{Institute of Neuroinformatics, University of Zurich and ETH Zurich, Zurich, Switzerland}
\affil[3]{Institute for Computer Graphics and Vision, Graz University of Technology, Graz, Austria}
\affil[4]{Ludwig Boltzmann Institute for Clinical Forensic Imaging, Graz, Austria}
\affil[5]{BioTechMed-Graz, Graz, Austria}
\affil[*]{\href{mailto:philipp.kainz@medunigraz.at}{\texttt{philipp.kainz@medunigraz.at}}}

\date{}
\flushbottom
\maketitle
\thispagestyle{empty}

\begin{abstract}
Segmentation of histopathology sections is an ubiquitous requirement in digital pathology and due to the large variability of biological tissue, machine learning techniques have shown superior performance over standard image processing methods. 
As part of the \challenge\ colon gland segmentation challenge, we present a learning-based algorithm to segment glands in tissue of benign and malignant colorectal cancer. 
Images are preprocessed according to the Hematoxylin-Eosin staining protocol and two deep convolutional neural networks (CNN) are trained as pixel classifiers. 
The CNN predictions are then regularized using a figure-ground segmentation based on weighted total variation to produce the final segmentation result. 
On two test sets, our approach achieves a tissue classification accuracy of $98\%$ and $94\%$, making use of the inherent capability of our system to distinguish between benign and malignant tissue.

\end{abstract}

\section{Introduction}
\label{sec:introduction}
The variability of glandular structures in biological tissue poses a challenge to automated analysis of histopathology slides.
It has become a key requirement to quantitative morphology assessment and supporting cancer grading.  
Considering non-pathological cases only, automated segmentation algorithms must already be able to deal with significant variability in shape, size, location, texture and staining of glands.
Moreover, in pathological cases gland objects can tremendously differ from non-patho\-logical and benign glands, which further exacerbates finding a general solution to the segmentation problem.

Previous work on gland segmentation in colon tissue has used graphical models~\cite{Gunduz-Demir2010,Tosun2011,Sirinukunwattana2015a} or textural features~\cite{Farjam2007}.
Others worked on segmentation in prostatic cancer tissue using an integrated low-, high-level and contextual segmentation model~\cite{Naik2008}, probabilistic Markov models~\cite{Monaco2010}, k-means clustering and region growing~\cite{Peng2011}, spatial association of nuclei to gland lumen~\cite{Nguyen2012,Rashid2013}.
The reader is referred to the work of Sirinukunwattana~\textit{et al.}~\cite{Sirinukunwattana2015a} for a more detailed description of work related to glandular structure segmentation.
Deep learning methods, especially convolutional neural networks (CNNs)~\cite{LeCun2010}, have found applications in biomedical image analysis for different tasks: semantic segmentation~\cite{Pang2010}, mitosis detection~\cite{Ciresan2013} and classification~\cite{Malon2013}, and blood cell counting~\cite{Habibzadeh2013}.

\begin{figure}[!ht]
\ifnofigures
\else
\centering
\begin{tabular}{cc}
\includegraphics[width=0.45\textwidth]{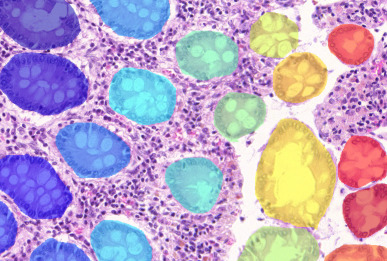} & 
\includegraphics[width=0.45\textwidth]{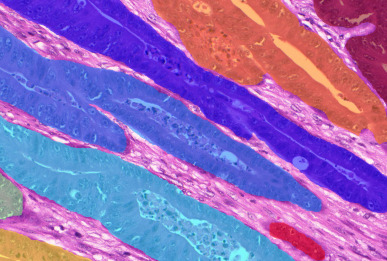} \\
\includegraphics[width=0.45\textwidth]{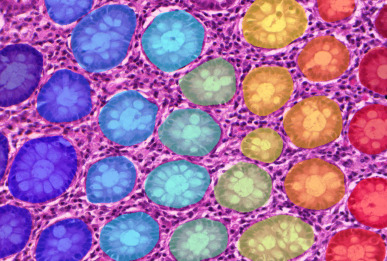} & 
\includegraphics[width=0.45\textwidth]{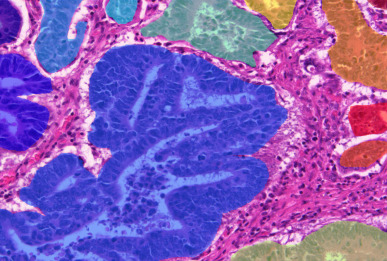} \\
\includegraphics[width=0.45\textwidth]{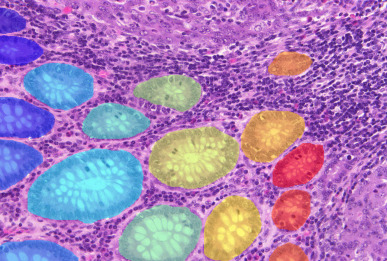} & 
\includegraphics[width=0.45\textwidth]{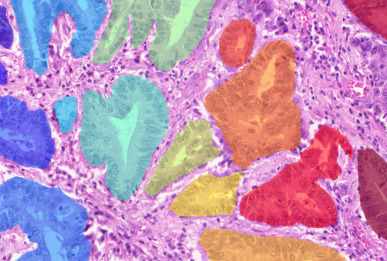} \\
\includegraphics[width=0.45\textwidth]{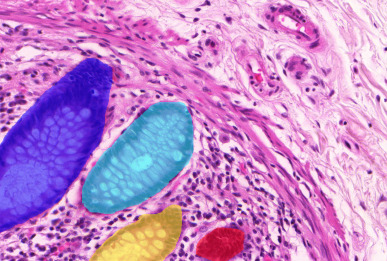} & 
\includegraphics[width=0.45\textwidth]{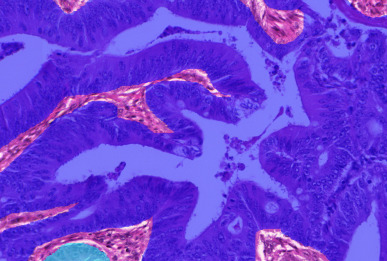} \\
(a) & (b) \\
\end{tabular}
\fi
\caption{\label{fig:datasetsamples}{\bf Samples of (a) benign and (b) malignant colorectal cancer sections in the \textit{Warwick-QU} dataset.}  
Ground truth labels in each image are available for each pixel and overlaid in different colors for individual objects.}
\end{figure}

In this work, we propose a learning-based strategy to semantically segment glands in the \textit{Warwick-QU} dataset, presented at the \challenge\ challenge\footnote{\href{http://www2.warwick.ac.uk/fac/sci/dcs/research/combi/research/bic/glascontest/}{\texttt{http://www2.warwick.ac.uk/fac/sci/dcs/research/combi/research/bic/glascontest/}}}. 
It contains 161 annotated images of benign and malignant colorectal adenocarcinoma, stained with Hematoxylin-Eosin (H\&E) and scanned at $20\times$ magnification. 
Fig.~\ref{fig:datasetsamples} shows some example images and their ground truth annotation. 
In each image, individual objects are annotated with the same label, illustrated by the different colors. 
To the challenge participants, information on whether an image shows benign or malignant tissue is only available in the training dataset. 
Three datasets were released during the contest and the total number of non-overlapping images (benign/malignant) in the training set, test set A and test set B is $85 (37/48)$, $60 (33/27)$, and $16 (12/4)$, respectively. 
These datasets further contained $795$, $666$, and $91$ individual glands.

The contributions of our work are twofold: (i) we present a novel deep learning scheme to generate classifier predictions for malignant and benign object and background pixels accompanied by a dedicated gland-separating refinement classifier that is able to distinguish touching objects, which pose a challenge for later segmentation. 
(ii) We use these classification results as the input for a simple, yet effective, globally optimal figure-ground segmentation approach based on a convex geodesic active contour formulation that regularizes the classifier predictions according to a minimal contour-length principle. 
Both technological contributions are described in section~\ref{sec:methods}, while the subsequent sections show and discuss the results of our novel approach applied to the datasets of the \challenge\ challenge.

\section{Methods}
\label{sec:methods}
We present a segmentation method for Hematoxylin-Eosin (H\&E) stained histopathological sections that proceeds in three steps: 
The raw RGB images are preprocessed to extract a robust representation of the tissue structure.
Subsequently, two classifiers are trained to predict glands (\textit{Object-Net}) and gland-separating structures (\textit{Separator-Net}) from the image. 
Finally, the outputs of the classifiers are combined and a figure-ground segmentation based on weighted total variation is used to produce the segmentation result. 

\subsection{Preprocessing H\&E Slides}
\label{subsec:preprocessing}
\begin{figure}[!ht]
\ifnofigures
\else
\centering
\includegraphics[width=1.0\textwidth]{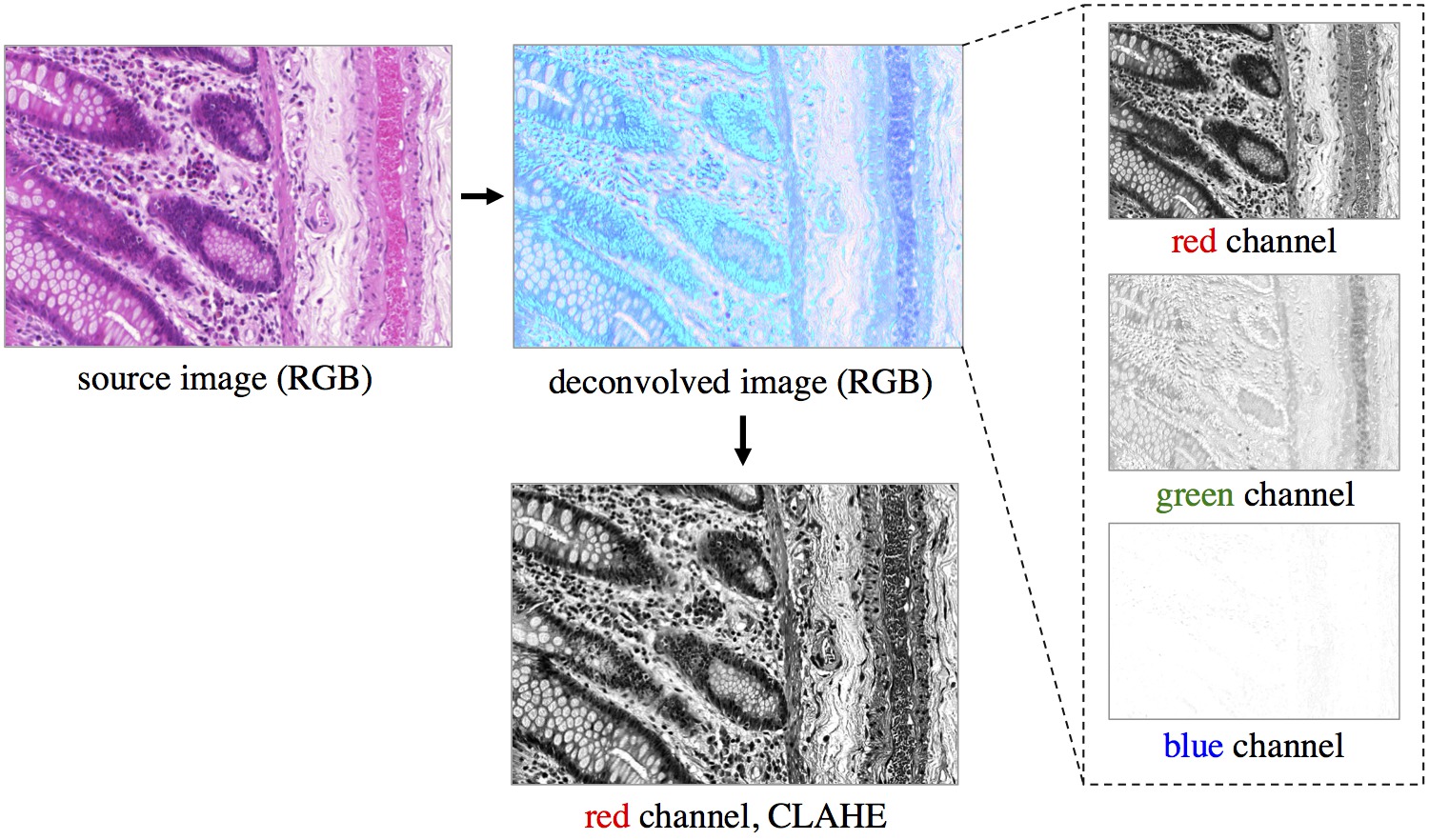}
\fi
\caption{\label{fig:preproc}{\bf Preprocessing of the RGB images.} 
Color deconvolution~\cite{Ruifrok2001} separates the Hematoxylin-Eosin stained tissue components. 
The red channel of the deconvolved image is processed by CLAHE~\cite{Zuiderveld1994} and taken as input to the pixel classifiers.}
\end{figure}

Prior to classification, the RGB images are preprocessed as shown in Fig.~\ref{fig:preproc}.
A standard color deconvolution~\cite{Ruifrok2001} is performed for the specific H\&E staining used in the provided dataset\footnote{We used the \textit{H\&E 2} setting in the implementation of G. Landini, available in Fiji~\cite{Schindelin2012}.}. 
It separates tissue components according to their staining, emphasizes the structure and inherently performs data whitening. 
The first (red) channel of the deconvolved RGB image contains most of the tissue structure information, so the other channels can be omitted.
In order to account for different staining contrasts and lighting conditions during image acquisition, contrast limited adaptive histogram equalization (CLAHE)~\cite{Zuiderveld1994} is applied.

\subsection{Learning Pixel Classifiers}
\label{subsec:learningpxclassifiers}
Given the large variability of both benign and malignant tissue in the \textit{Warwick-QU} dataset, we opted for CNNs due to their recently shown convincing performance in pixelwise classification of histopathology images~\cite{Ciresan2013} and to learn a rich set of features directly from images. 

\begin{figure}[!hb]
\ifnofigures
\else
\begin{tabular}{c}
\includegraphics[width=1.0\textwidth]{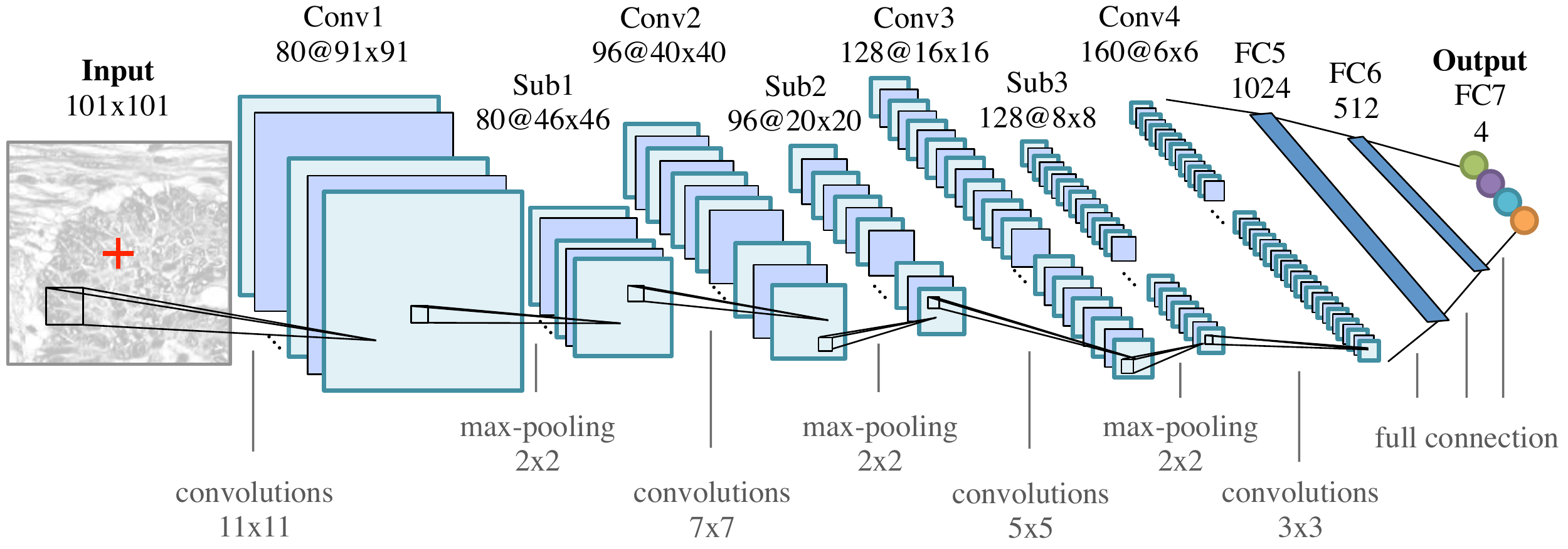} \\
(a) \textit{Object-Net} architecture\\
\\
\includegraphics[width=1.0\textwidth]{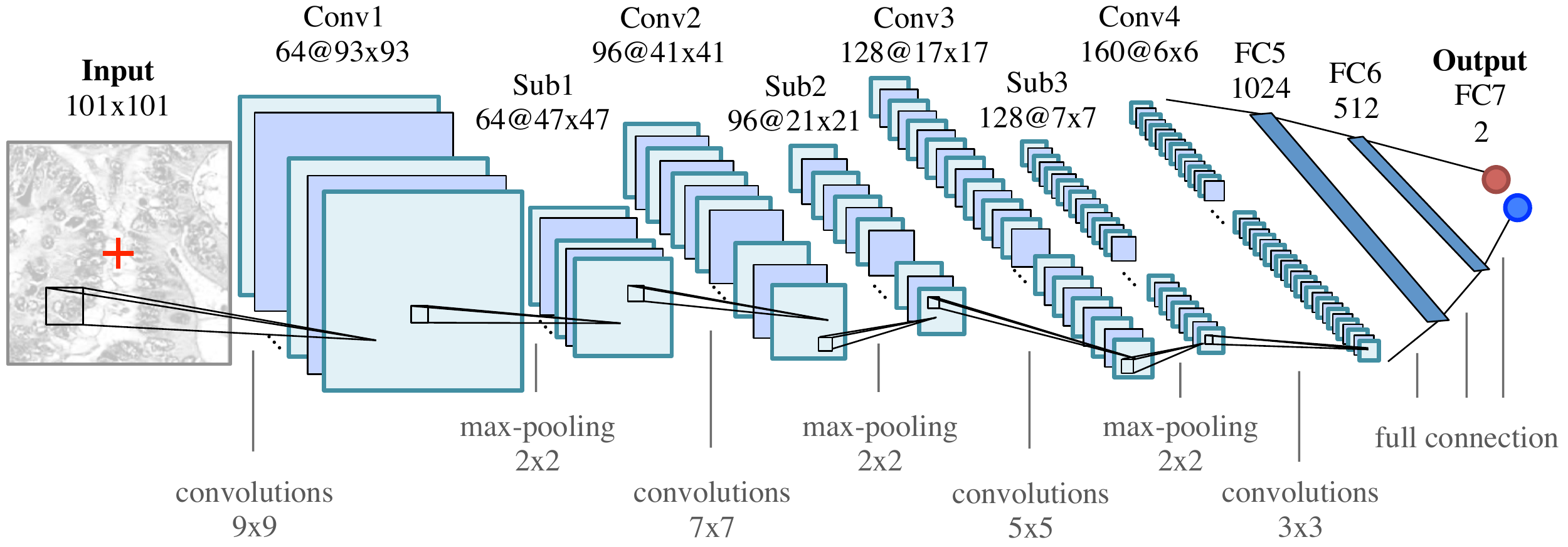} \\
(b) \textit{Separator-Net} architecture
\end{tabular}
\fi
\caption{\label{fig:architectures}{\bf CNN classifier architectures of (a) the \textit{Object-Net} and (b) the \textit{Separator-Net}.} 
Both architectures have $K=7$ ($k=1,\ldots,K$) layers, are identical in the number of convolutional (Conv$k$), max-pooling (Sub$k$), and fully connected (FC$k$) layers, but differ in convolution kernel size, size and number of the feature maps, as well as number of output units. 
The probability distribution over $L$ labels of the center pixel $\mathbf{x}=(u,v)^\top$ (marked as red cross in the input patch) is predicted by the CNNs.}
\end{figure}

The general architecture of both CNNs is motivated by a classical LeNet-5 architecture~\cite{LeCun1998} and consists of $K=7$ ($k=1,\ldots,K$) layers: four convolutional layers (Conv$k$) for feature learning and three fully connected (FC$k$) layers as feature classifier, see Fig.~\ref{fig:architectures}.
The rectified linear unit (ReLU) nonlinearity ($f(x)=\max(0,x)$) is used as the activation function throughout all layers in the networks. 
All convolutional layers consist of a set of learnable square 2D filters with pixel stride 1, followed by ReLU activation. 
Subsampling (max-pooling) layers (Sub$k$, $2\times 2$), accounting for translation invariance, are used after the first three convolutional layers and are counted as part of the convolutional layer. 
The final pixelwise classification of an input image is obtained by sliding a window over the image, and classifying the center pixel of that window. 

For training minibatch stochastic gradient descent (MBSGD) with momentum, weight decay, and dropout regularization is used to minimize a negative log-likelihood loss function.

\subsubsection{\textit{Object-Net}: Classifying Gland Objects}
\label{subsubsec:objectnet}
\begin{figure}[!ht]
\ifnofigures
\else
\centering
\begin{tabular}{p{0.3\textwidth}p{0.3\textwidth}p{0.3\textwidth}}
\multicolumn{3}{c}{\includegraphics[width=1.0\textwidth]{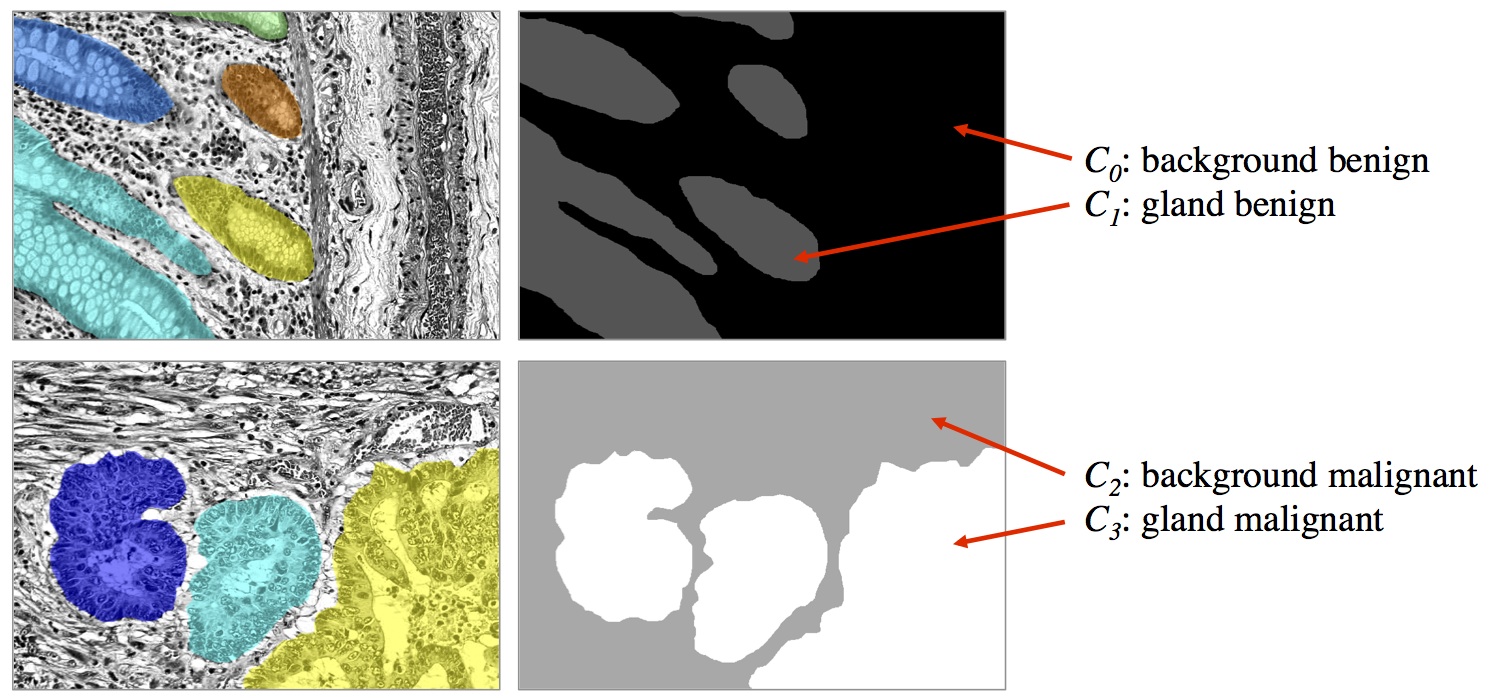}
}\\
\hspace*{2.2cm}(a) & \hspace*{2.2cm}(b) & \\ 
\end{tabular}
\fi
\caption{\label{fig:label_transform_object_net}{\bf Ground truth transformation for learning the four-class classification on the preprocessed images with the \textit{Object-Net}.} 
The first row shows a benign case, the second row shows a malignant case.
(a) Preprocessed images with overlaid individual ground truth object annotations. 
(b) Provided annotations were transformed into four labels for benign background ($C_0$), benign gland ($C_1$), malignant background ($C_2$) and malignant gland ($C_3$).
}
\end{figure}

The goal of the \textit{Object-Net} is to predict the probability of a pixel belonging to a gland or background.  
One could now define a binary classification problem, but malignant and benign tissue express express unique features, which are not found in the other tissue type, and which can thus complicate the learning problem.
We therefore formulate an alternative, four-class classification problem, in which we distinguish ($L=4$, with $l=0,...,L-1$): background benign ($C_0$), gland benign ($C_1$), background malignant ($C_2$), and gland malignant ($C_3$). 
In order to do that it is necessary to transform the provided ground truth labels to reflect benignity and malignancy as well. 
The annotation images are binarized and a new label is assigned to pixels belonging to each class $C_l$, see Fig.~\ref{fig:label_transform_object_net}. 

The input to the CNN is an image patch $I(\mathbf{x})$ of size $101\times101$ pixels, centered at an image location $\mathbf{x}=(u,v)^\top$, where $\mathbf{x}\in\Omega$ and $\Omega$ denotes the image domain. 
A given patch $I(\mathbf{x})$ is convolved with 80 filters ($11\times11$) in the first convolutional layer, in the second layer with 96 filters ($7\times7$), in the third layer with 128 filters ($5\times5$), and in the last layer with 160 filters ($3\times3$), see Fig.~\ref{fig:architectures}(a). 
The three subsequent fully connected layers FC5-FC7 of the classifier contain 1024, 512, and four output units, respectively. 
The output of FC7 is fed into a softmax function, producing the center pixel's probability distribution over the labels. 
The probability for each class $l$ is stored in a corresponding map $I_{C_{l}}(\mathbf{x})$.

\subsubsection{\textit{Separator-Net}: Classifying Gland-separating Structures}
\label{subsubsec:separatornet}
Initial experiments have shown that taking pixelwise predictions only from the \textit{Object-Net} were insufficient in order to separate very close gland objects.  
Hence, a second CNN, the \textit{Separator-Net}, is trained to predict structures in the image that are separating such objects.
This learning problem is formulated as binary classification task. 

As depicted in Fig.~\ref{fig:architectures}(b), the CNN structure is similar to the \textit{Object-Net}: a given input image patch $I(\mathbf{x})$ of size $101\times101$ pixels is convolved with 64 filters ($9\times9$) in the first convolutional layer, in the second layer with 96 filters ($7\times7$), in the third layer with 128 filters ($5\times5$), and in the last layer with 160 filters ($3\times3$). 
The three subsequent fully connected layers FC5-FC7 of the classifier contain 1024, 512, and two output units, respectively. 
The output of the last layer (FC7) is fed into a softmax function to produce the probability distribution over the labels for the center pixel. 
The probability for a pixel $\mathbf{x}$ belonging to a gland-separating structure is stored in the corresponding probability map $S(\mathbf{x})$.

\subsubsection{Refining CNN Outputs}
\label{subsubsec:combiningCNNoutputs}
Once all probability maps have been obtained, the \textit{Object-Net} predictions $I_{C_{l}}(\mathbf{x})$ are refined with the \textit{Separator-Net} predictions $S(\mathbf{x})$ to emphasize the gland borders and prevent merging of close objects. 
The subsequent figure-ground segmentation algorithm requires a single foreground and background map to produce the final segmentation result, so outputs are combined as follows.

The foreground probability map $p_{fg}$ is constructed by
\begin{equation}
\label{eq:pfg}
p_{fg}(\mathbf{x}) = \max\left\lbrace\left(\sum_{l\in\lbrace 1,3\rbrace}I_{C_{l}}(\mathbf{x})\right) - \rho S(\mathbf{x}), 0\right\rbrace, 
\end{equation}
\noindent where $\rho\in\left[ 0,1\right]$ controls the influence of the refinements done by the separator predictions.
Similarly, evaluating Eq.~\eqref{eq:pbg} produces the background probability map: 
\begin{equation}
\label{eq:pbg}
p_{bg}(\mathbf{x}) = \min\left\lbrace\left(\sum_{l\in\lbrace 0,2\rbrace}I_{C_{l}}(\mathbf{x})\right) + \rho S(\mathbf{x}), 1\right\rbrace.
\end{equation}

\subsection{Total Variation Segmentation}
\label{subsec:TVseg}
To generate a final segmentation, the following continuous non-smooth energy functional $E_{seg}(u)$~\cite{reinbacher10,Hammernik2015} is minimized:
\begin{equation}
\begin{gathered}
\MIN{u} E_{seg}(u) = \MIN{u} \intOmega g(\mathbf{x})\TVNormCont{u(\mathbf{x})} \dd{\mathbf{x}} + \lambda \intOmega u(\mathbf{x})\cdot w(\mathbf{x}) \dd{\mathbf{x}} \\ \suchThat u \in C_{box}=\lbrace u: u(\mathbf{x}) \in [0,1],\,\forall\,\mathbf{x}\in\Omega \rbrace
\label{eq:energy}
\end{gathered}
\end{equation}

\noindent where $\Omega$ denotes the image domain and $u \in C^1\,:\,\Omega \mapsto \R$ is smooth. 
The first term denotes the $g$-weighted total variation (TV) semi-norm which is a reformulation of the geodesic active contour energy~\cite{Bresson2007}. The edge function $g(\mathbf{x})$ is defined as
\begin{equation}
\label{eq:edge}
g (\mathbf{x})= e^{-\alpha\norm{\nabla I(\mathbf{x})}^{\beta}},\, \alpha, \beta > 0,
\end{equation}

\noindent where $\nabla I(\mathbf{x})$ is the gradient of the input image, thus attracting the segmentation towards large gradients.
The second term in Eq.~\eqref{eq:energy} is the data term with $w$ describing a weighting map. 
The values in $w$ have to be chosen negative if $u$ should be foreground and positive if $u$ should be background. 
If values in $w$ are set to zero, the pure weighted TV energy is minimized seeking for a minimal contour length segmentation. 
We use the refined outputs from the previous classification step (Eqs.~\eqref{eq:pfg} and~\eqref{eq:pbg}) and introduce a threshold $\tau$ to ensure a minimum class confidence in a map $p$:

\begin{equation}
\label{eq:p_x}
p(\mathbf{x}) = \begin{cases} 	0    		  & \textIf{p(\mathbf{x}) < \tau} \\
                  				w(\mathbf{x}) & \mathrm{otherwise}				\end{cases}.
\end{equation}

\noindent The weighting map $w$ is derived by applying the logit transformation: 

\begin{equation}
\label{eq:w_x}
w (\mathbf{x}) = \begin{cases} -(log(p_{fg}(\mathbf{x})) - log(1 - p_{fg}(\mathbf{x}))) & \textIf{p_{fg}(\mathbf{x}) > p_{bg}(\mathbf{x})} \\
                      log(p_{bg}(\mathbf{x})) - log(1 - p_{bg}(\mathbf{x})) & \textIf{p_{fg}(\mathbf{x}) \le p_{bg}(\mathbf{x})}\end{cases}.
\end{equation}

\noindent The regularization parameter $\lambda$ defines the trade-off between our data term and the weighted TV semi-norm. 
The stated convex problem in Eq.~\eqref{eq:energy} can be solved for its global optimum efficiently using the primal-dual algorithm~\cite{chambolle11pdAlgo}, which can be implemented very efficiently using NVidia CUDA, thus making use of the parallel computing power of recent GPUs.
As the segmentation $u$ is continuous, the final segmentation is achieved by thresholding $u$ with a value of $0.5$.
We optimize the free parameters $\alpha$, $\beta$ and $\lambda$ by performing a grid search in a suitable range of these values ($\alpha \in [0.5, 15]$, $\beta \in [0.35, 0.95]$ and $\lambda \in [0.01, 10]$), where all 85 annotated training images are used to tune these parameters based on the Dice coefficient.

\subsection{Implementation Details}
\label{subsec:implementationdetails}
\subsubsection{Training Dataset Sampling}
\label{subsubsec:datasetsampling}
For the sake of execution speed when using a sliding window approach, the images are rescaled to half resolution prior to classification and upsampled with bilinear interpolation to their original size afterwards. 
The size of the input patch $I(\mathbf{x})$ is chosen to be $101\times101$ pixels, such that sufficient contextual information is available to classify the center pixel. 

\begin{figure}[!htb]
\ifnofigures
\else
\centering
\begin{tabular}{cc}
\includegraphics[width=0.48\textwidth]{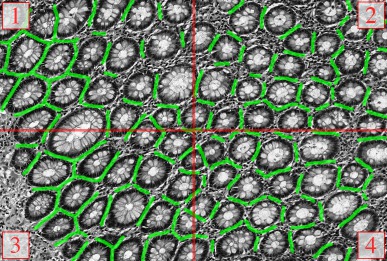} & 
\includegraphics[width=0.48\textwidth]{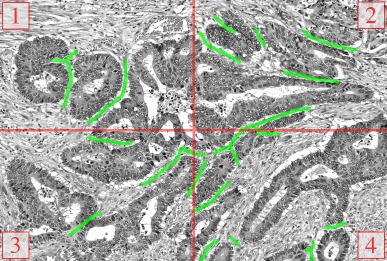}\\
\end{tabular}
\fi
\caption{\label{fig:sepnetgroundtruth}
{\bf Manual ground truth annotations for gland-separating structures.}
Stitched images from four tiles (numbers in red boxes), red lines denote the tile borders.
Manual annotations of pixels belonging to gland-separating structures are shown as green lines, the thickness of lines is increased for better illustration.}
\end{figure}

The majority of training images (79) have a size of $775\times522$ pixels, and resizing reduces them to $387\times261$ pixels. 
If we just considered the valid part without border extension for sampling the patches for the training dataset, we would actually lose approximately $46\%$ of the labeled pixels when using a patch size of $101\times101$ pixels. 
On the other hand, we would introduce a significant number of boundary artifacts by artificially extending the border to make use of all labeled pixels. 
Fortunately, most images are tiles of a bigger image and can thus be stitched seamlessly to obtain a total of 19 images\footnote{In one case, stitching was not possible, since only 3 tiles were available. These 3 tiles, and the remaining 6 images, that were not part of a bigger image, were treated as individual images.} (Fig.~\ref{fig:sepnetgroundtruth}), where we can sample enough patches without heavily relying on artificial border extension. 

In principle, we pursued the same sampling strategy for the \textit{Separator-Net}, but were required to create the ground truth labels manually. 
We annotated all pixels that belong to a structure very close to two or more gland borders. 
The green lines in Fig.~\ref{fig:sepnetgroundtruth} illustrate the additional manual annotation of the separating structures. 
Due to the low number of foreground samples when compared to the \textit{Object-Net}, the number of foreground samples for the \textit{Separator-Net} was artificially increased by exploiting the problem's requirement for rotation-invariance and adding nine additional rotated versions of the patch, i.e. every $36^{\circ}$. 

\subsubsection{CNN Training}
\label{subsec:cnntraining}
\begin{figure}[!htb]
\ifnofigures
\else
\centering
\includegraphics[width=0.5\textwidth]{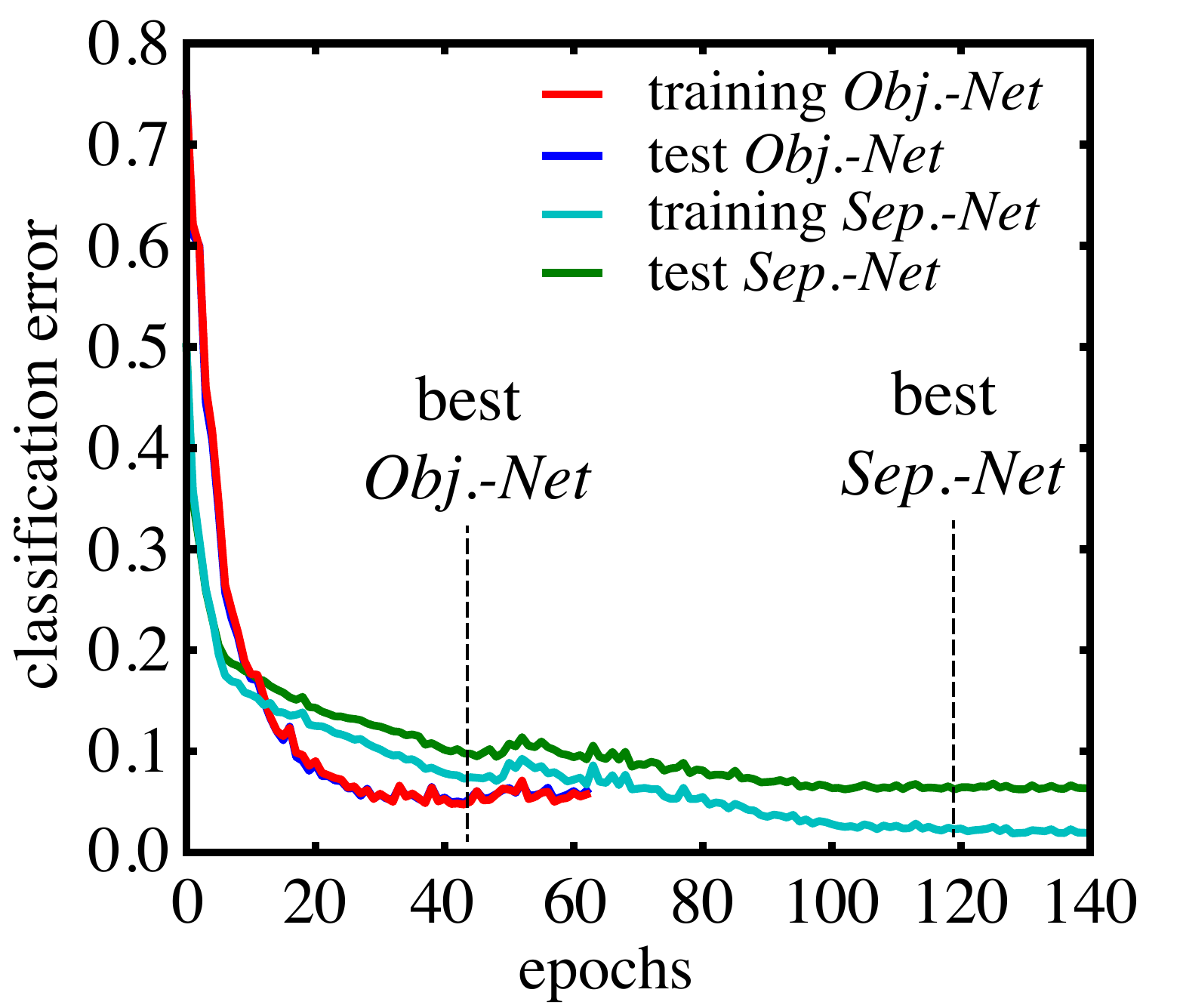}
\fi
\caption{\label{fig:errors}{\bf CNN training progress.}
Classification error over epochs on a subset of the training data (training error), and on the held-out test set. 
The \textit{Object-Net} reaches below $4.9\%$ test error after 43 epochs, the \textit{Seperator-Net} reaches $6.2\%$ test error after 119 epochs.
}
\end{figure}

Both CNNs were trained on a balanced training set of $125,000$ image patches per class.  Patches in the training sets were sampled at random from the available pool of training images. 
Training and test sets reflect approximately the same distribution of samples over images. 
The size of the minibatches in the MBSGD was set to $200$ samples and the networks were trained until the stopping criterion was met: no further improvement of the error rate on a held-out test set over 20 epochs. 
We set the initial learning rate $\eta_0=0.0025$, with a linear decay saturating at $0.2\eta_0$ after 100 epochs.  
For all layers, a weight decay was chosen to be $0.005$ and the dropout rate was set to $0.5$.
We used an adaptive momentum term starting at $0.8$ and increasing to $0.99$ after 50 epochs, such that with progressing training the updates are influenced by a larger number of samples than at the beginning. 

\begin{figure}[!htb]
\ifnofigures
\else
\centering
\begin{tabular}{cc}
\includegraphics[scale=1.25]{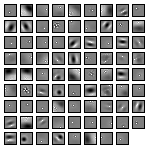} & 
\includegraphics[scale=1.25]{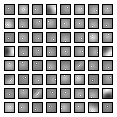}\\
(a) & (b) \\
\end{tabular}
\fi
\caption{\label{fig:filters}{\bf CNN training results.}  
(a) 80 $11\times11$ filters of the first layer in \textit{Object-Net} and (b) 64 $9\times9$ filters in the \textit{Separator-Net}.}
\end{figure}

Fig.~\ref{fig:errors} shows the classification error rate as a function of the training duration in epochs. 
Each class was represented with $5,000$ samples in the test set for the \textit{Object-Net}, and $10,000$ for the \textit{Separator-Net}, respectively. 
The training error is actually estimated on a fixed subset of the training data ($20,000$ samples), to get an intuition when overfitting starts.
The \textit{Object-Net} achieves the best performance after $43$ epochs, with a minimum training error of $0.0475$ and a minimum test error of $0.0492$. 
Training of the \textit{Separator-Net} continued until the lowest training error of $0.0231$ and test error of $0.0624$ was reached after $119$ epochs. 
Fig.~\ref{fig:filters} shows the learned filters of the first convolutional layer in both networks.
The CNN models were implemented in Pylearn2~\cite{Goodfellow2013}, a machine learning library built on top of Theano~\cite{Bergstra2010,Bastien2012}. 

\section{Results}
\label{sec:results}
\subsection{Colon Gland Segmentation}
\label{subsec:colonglandsegmentation}
The grid search resulted in $\alpha=10$, $\beta=0.95$ and $\lambda=0.1$ as parameters optimizing the TV segmentation based on the Dice score. 
The confidence threshold for foreground and background was determined empirically and fixed to $\tau=0.65$.
Separator predictions were fully considered for refining the \textit{Object-Net} predictions ($\rho=1$).
\begin{table}[!htb]
\centering
\caption{\label{tab:segmentationresults}{\bf Segmentation performance metrics for the \textit{Warwick-QU} dataset used in the \challenge\ challenge.}}
\begin{small}
\begin{tabular}{cccccc}
\toprule
\textbf{Dataset} &\textbf{Precision} & \textbf{Recall} & \textbf{F1-score} & \textbf{Object-Dice} & \textbf{Hausdorff} \\
\midrule
\multicolumn{6}{c}{\textit{without} separator refinement}\\
\midrule
Training & \textbf{0.97}(0.09) & 0.67(0.21) & 0.78(0.17) & 0.81(0.16) & 116.89(115.18) \\
Test A & \textbf{0.83}(0.22) & 0.60(0.24) & 0.67(0.20) & 0.70(0.15) & 137.44(78.53) \\
Test B & \textbf{0.70}(0.35) & 0.48(0.30) & 0.50(0.26) & 0.58(0.19) & 249.37(114.69) \\
\midrule
\multicolumn{6}{c}{\textit{with} separator refinement}\\
\midrule
Training & 0.91(0.15) & \textbf{0.85}(0.14) & \textbf{0.87}(0.12) & \textbf{0.88}(0.09) & \textbf{61.36}(61.36) \\
Test A & 0.67(0.24) & \textbf{0.77(}0.22) & \textbf{0.68}(0.20) & \textbf{0.75}(0.13) & \textbf{103.49}(72.38) \\
Test B & 0.51(0.30) & \textbf{0.70}(0.32) & \textbf{0.55}(0.28) & \textbf{0.61}(0.22) & \textbf{213.58}(119.15) \\
\bottomrule
\end{tabular}
\begin{flushleft} 
Metrics are reported as mean and standard deviation, best results are printed in bold.
Performance on the training set is reported on all $85$ training images.
Test set A consists of $60$ images, test set B of $16$ images. 
Except for values of the Hausdorff distance, higher values are superior.
\end{flushleft}
\end{small}
\end{table}

In Table~\ref{tab:segmentationresults}, we report performance metrics\footnote{The evaluation scripts were kindly provided by the contest organizers and are available from \href{http://www2.warwick.ac.uk/fac/sci/dcs/research/combi/research/bic/glascontest/evaluation/}{\texttt{http://www2.warwick.ac.uk/fac/sci/dcs/research/combi/research/bic/glascontest/ evaluation/}}.} for detection (precision, recall, F1-score), segmentation (object-level Dice), and shape (Hausdorff distance) on the training set, as well as test set A and B as mean and standard deviation (SD). 
Blobs with an area less than $500$ pixels were removed and all remaining blobs were labeled with unique identifiers before computing the measures.

\begin{figure}[!ht]
\ifnofigures
\else
\centering
\begin{tabular}{ccc}
\includegraphics[width=0.3\linewidth]{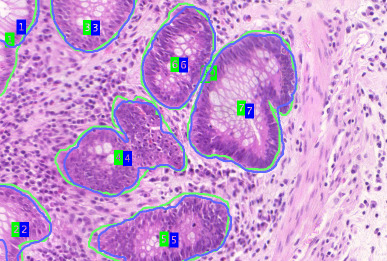} &
\includegraphics[width=0.3\linewidth]{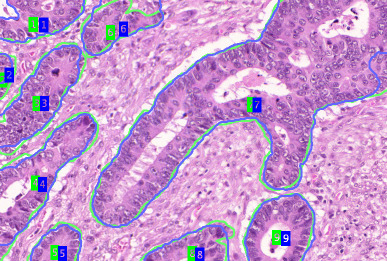} &
\includegraphics[width=0.3\linewidth]{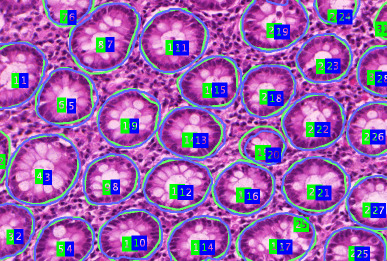}
 \\
\includegraphics[width=0.3\linewidth]{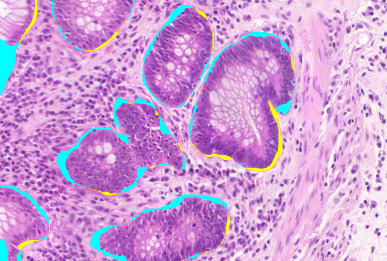} & 
\includegraphics[width=0.3\linewidth]{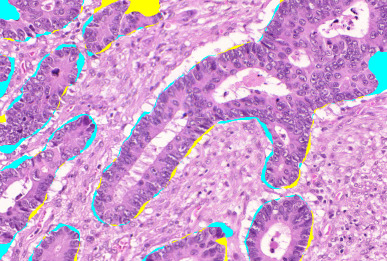} &
\includegraphics[width=0.3\linewidth]{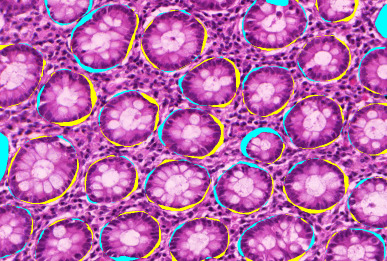} 
 \\
(a) benign & (b) malignant & (c) benign \\[0.2cm]
\includegraphics[width=0.3\linewidth]{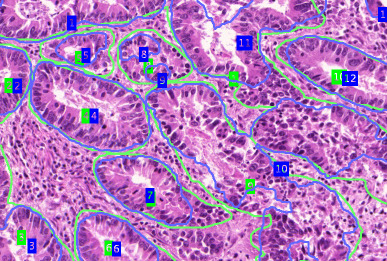} &
\includegraphics[width=0.3\linewidth]{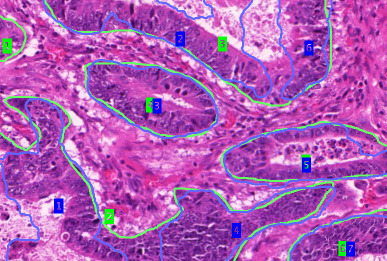} &
\includegraphics[width=0.3\linewidth]{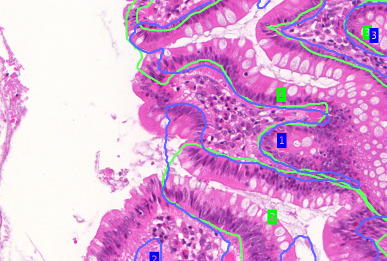}  
\\
\includegraphics[width=0.3\linewidth]{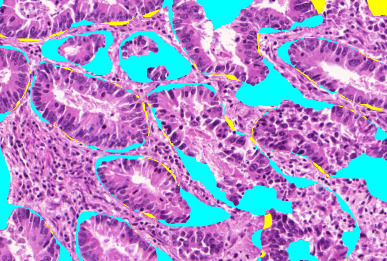} &
\includegraphics[width=0.3\linewidth]{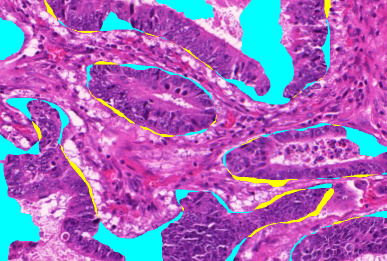} &
\includegraphics[width=0.3\linewidth]{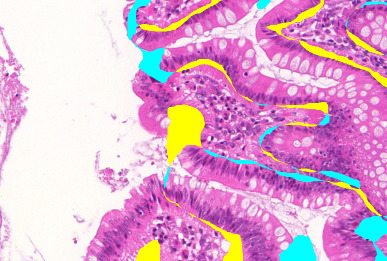}  
\\
(d) malignant & (e) malignant & (f) benign \\
\end{tabular}
\fi
\caption{\label{fig:qualres_train}{\bf Qualitative segmentation results on images of the training dataset.} 
Even rows show the outline of ground truth in green and the segmentation result in blue. 
The numbers refer to the unique objects within the image.
Odd rows show the segmentation difference: false negative pixels are colored in cyan, and false positives are colored in yellow. 
(a-c) show examples, where our segmentation algorithm works well, (d-f) show different types of segmentation errors.
}
\end{figure}

\begin{figure}[!htb]
\ifnofigures
\else
\centering
\begin{tabular}{ccc}
\includegraphics[width=0.3\linewidth]{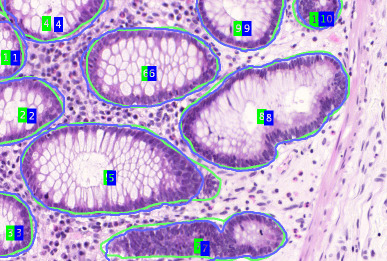} & 
\includegraphics[width=0.3\linewidth]{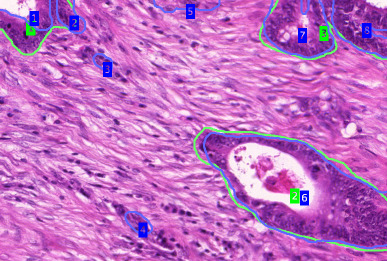} &
\includegraphics[width=0.3\linewidth]{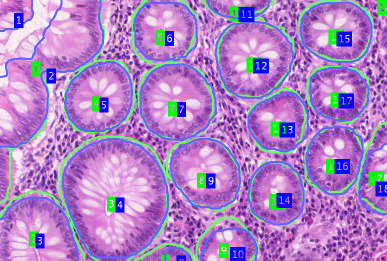} 
 \\
\includegraphics[width=0.3\linewidth]{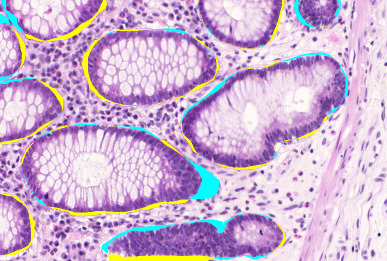} & 
\includegraphics[width=0.3\linewidth]{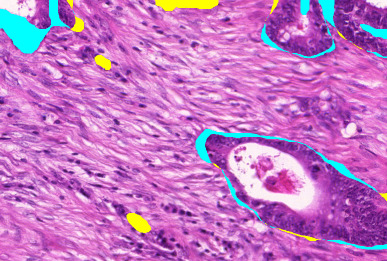} &
\includegraphics[width=0.3\linewidth]{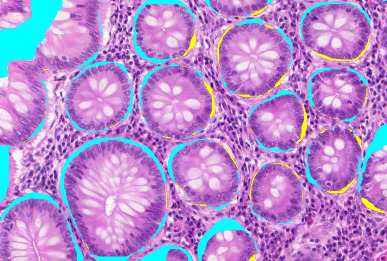} 
\\
(a) benign & (b) malignant & (c) benign \\[0.2cm]
\includegraphics[width=0.3\linewidth]{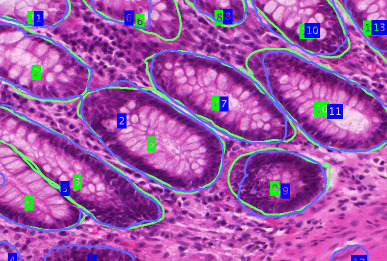} & 
\includegraphics[width=0.3\linewidth]{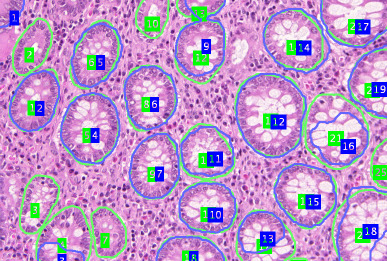} &
\includegraphics[width=0.3\linewidth]{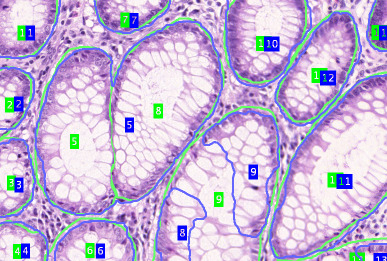} 
\\
\includegraphics[width=0.3\linewidth]{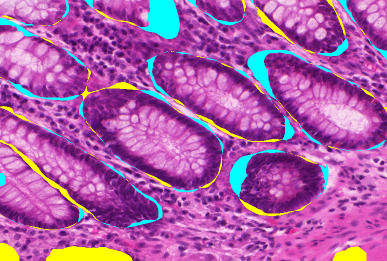} & 
\includegraphics[width=0.3\linewidth]{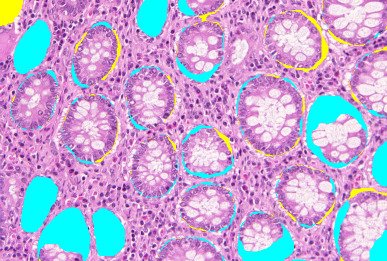} &
\includegraphics[width=0.3\linewidth]{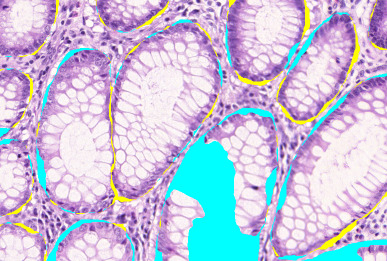} 
\\
(d) benign & (e) benign & (f) malignant \\
\end{tabular}
\fi
\caption{\label{fig:qualres_testA}{\bf Qualitative segmentation results on images of test dataset A.}
Even rows show the segmentation (blue outline) and ground truth (green outline), odd rows show the differences, where false negative pixels are cyan, and false positive pixels are yellow.
(a-c) show reasonable segmentation results, in (d-f) different segmentation errors are shown.
}
\end{figure}

\begin{figure}[!htb]
\ifnofigures
\else
\centering
\begin{tabular}{ccc}
\includegraphics[width=0.3\linewidth]{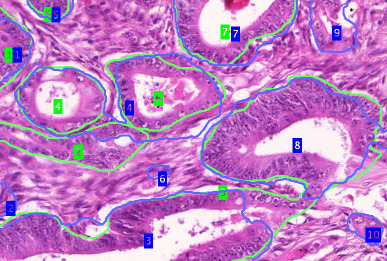} & 
\includegraphics[width=0.3\linewidth]{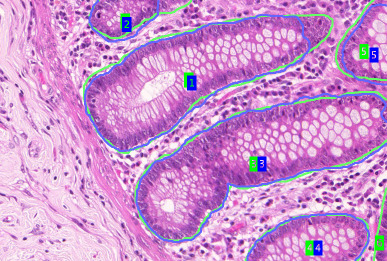} &
\includegraphics[width=0.3\linewidth]{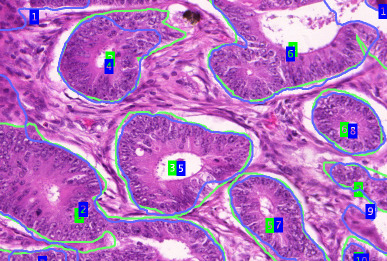} \\
\includegraphics[width=0.3\linewidth]{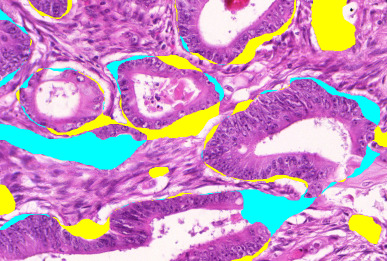} & 
\includegraphics[width=0.3\linewidth]{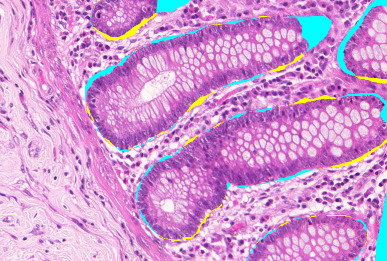} &
\includegraphics[width=0.3\linewidth]{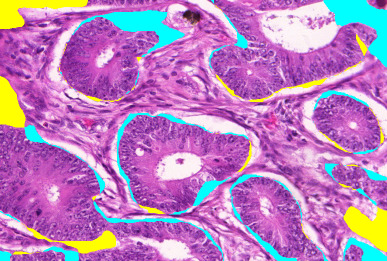} 
 \\
(a) malignant & (b) benign & (c) malignant \\[0.2cm]
\includegraphics[width=0.3\linewidth]{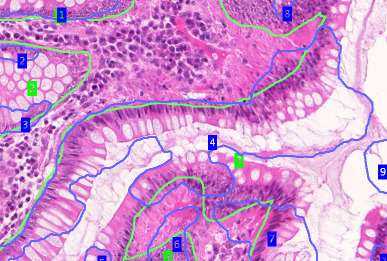} & 
\includegraphics[width=0.3\linewidth]{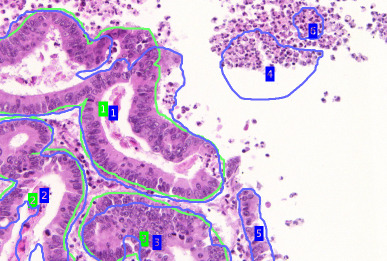} &
\includegraphics[width=0.3\linewidth]{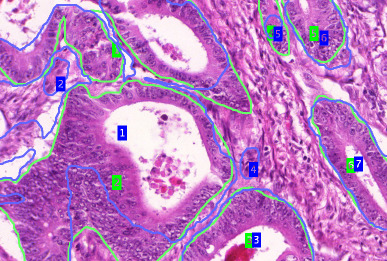} 
\\
\includegraphics[width=0.3\linewidth]{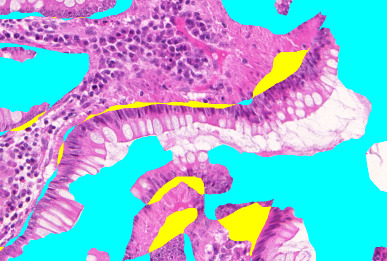} & 
\includegraphics[width=0.3\linewidth]{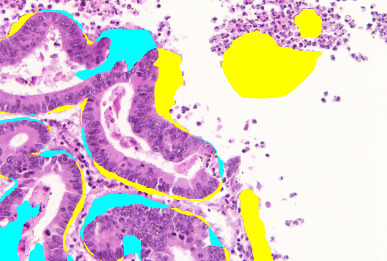} &
\includegraphics[width=0.3\linewidth]{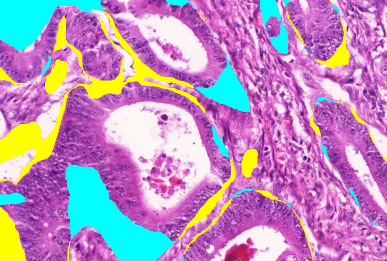} 
\\
(d) benign & (e) malignant & (f) malignant \\
\end{tabular}
\fi 
\caption{\label{fig:qualres_testB}{\bf Qualitative segmentation results on images of test dataset B.} 
Even rows show the segmentation (blue outline) and ground truth (green outline), odd rows show the differences, where false negative pixels are cyan, and false positive pixels are yellow.
(a-c) show reasonable segmentation results, in (d-f) different segmentation errors are shown.
}
\end{figure}

Compared to using predictions only from the \textit{Object-Net}, the segmentation performance improved with separator refinement. 
Malignant cases are harder to segment due to their irregular shape and pathological variations in the tissue. 
Fig.~\ref{fig:qualres_train} illustrates some qualitative example segmentation results on the training data set, Fig.~\ref{fig:qualres_testA} and Fig.~\ref{fig:qualres_testB} show results on test set A and B, respectively.

The average total runtime for segmenting a $577\times522$ image is 5 minutes using an NVidia GeForce Titan Black 6GB GPU.

\subsection{Benignity and Malignancy Classification}
In the proposed approach, the \textit{Object-Net} inherently learns a discrimination of benign ($c=0$) and malignant ($c=1$) tissue, since the labels for benign and malignant are available in the training dataset and we defined a four-class classification problem.
Instead of combining the probability maps for glands and background as done for segmentation, we combine the maps for benignity and malignancy.
Subsequently, the average probabilities for a benign case can be computed as

\begin{equation}
\label{eq:px_integrator}
\overline{\mathrm{P}}(c=0|I_{C_0},I_{C_1}) = \frac{1}{\vert\Omega\vert}\sum_{\mathbf{x}\in\Omega} I_{C_0}(\mathbf{x}) + I_{C_1}(\mathbf{x}),
\end{equation}

\noindent and for a malignant case as

\begin{equation}
\label{eq:px_integrator1}
\overline{\mathrm{P}}(c=1|I_{C_2},I_{C_3}) = \frac{1}{\vert\Omega\vert}\sum_{\mathbf{x}\in\Omega} I_{C_2}(\mathbf{x}) + I_{C_3}(\mathbf{x}),
\end{equation}

\noindent where $\vert\Omega\vert$ is the number of pixels in the image domain $\Omega$.
The maximum of both values finally indicates the prediction: 
\begin{equation}
\label{eq:ben_mal}
c^{*} = \argmax_c\left\lbrace \overline{\mathrm{P}}(c|\cdot,\cdot)\right\rbrace.
\end{equation}

We evaluated the classification performance for benign and malignant tissue on the two test sets A and B and achieved an accuracy of $98.33\%$ and $93.75\%$. 
The average (SD) decision confidence in test set A was $0.84(0.13)$ for benign and $0.81(0.11)$ for malignant, and in test set B $0.74(0.11)$ and $0.86(0.15)$, respectively.

\section{Discussion and Conclusions}
\label{sec:discussion}
This paper presented a method to segment glands in H\&E stained histopathological images of colorectal cancer using deep convolutional neural networks and total variation segmentation. 
As our main contribution, we showed that segmentation results can be greatly improved when the predictions of the \textit{Object-Net} are refined with the learned gland-separating structures of the \textit{Separator-Net}. 
Adding the separators does not only regulate the trade-off between precision and recall, but generally improves the performance scores for detection (F1-score), segmentation (Dice) and shape (Hausdorff). 
The final ranking as well as the test set performance results of other algorithms participating in this challenge are available online at the contest website\footnote{\href{http://www2.warwick.ac.uk/fac/sci/dcs/research/combi/research/bic/glascontest/results/}{\texttt{http://www2.warwick.ac.uk/fac/sci/dcs/research/combi/research/bic/glascontest/ results/}}}, which is continuously being updated by algorithms from new participating groups.

Our approach inherently allows to very accurately discriminate benign and malignant cases, because the \textit{Object-Net} was trained on labels for both cases. 
The average confidence for a decision towards benignity and malignancy is acceptable.
Nevertheless, we cannot distinguish more detailed histologic grades among these cases, since there was no information (e.g. high- or low-grade) available in addition to the segmentation ground truth. 

\section*{Acknowledgements}
The authors are grateful to the organizers of the \challenge\ challenge for providing (i) the \textit{Warwick-QU} image dataset, and (ii) the MATLAB evaluation scripts for computing performance measures that are comparable among the participating teams.
Further thanks goes to Julien Martel for fruitful discussions in early phases of this challenge.

{\small
\bibliographystyle{unsrt}
\bibliography{references}
}

\end{document}